\documentclass[10pt,journal,compsoc]{IEEEtran}
\usepackage[utf8]{inputenc} 
\usepackage[T1]{fontenc}    
\usepackage{url}            
\usepackage{booktabs}       
\usepackage{amsfonts}       
\usepackage{nicefrac}       
\usepackage{microtype}      
\usepackage{xcolor}         
\usepackage{graphicx}
\usepackage{amsmath}
\usepackage{amssymb}
\usepackage{algorithm}
\usepackage{algorithmic}
\usepackage{bm}
\usepackage{mathrsfs}
\usepackage{pifont}
\usepackage{ragged2e}
\usepackage{enumitem}
\usepackage{makecell}
\usepackage{colortbl}
\usepackage{rotating}
\usepackage[normalem]{ulem}
\usepackage{dcolumn}
\usepackage{multirow}
\usepackage{wrapfig}

\usepackage[pagebackref=true,breaklinks=true,colorlinks,bookmarks=false]{hyperref}

\graphicspath{{./Imgs/}}
\DeclareGraphicsExtensions{.pdf,.jpg,.png}

\def\ie{\emph{i.e.}}

\def\etal{{\em et al.~}}

\makeatletter
\newcommand{\thickhline}{%
    \noalign {\ifnum 0=`}\fi \hrule height 1pt
    \futurelet \reserved@a \@xhline
}

\newcommand{\myPara}[1]{\vspace{.12in}\noindent\textbf{#1}\quad}

\ifCLASSOPTIONcompsoc
  \usepackage[nocompress]{cite}
\else
  \usepackage{cite}
\fi
\ifCLASSINFOpdf
\else
\fi
\begin{document}
%
\title{Boosting Few-shot Semantic Segmentation with Transformers}


\author{ Guolei Sun,~Yun Liu,~Jingyun Liang,~Luc Van Gool
\IEEEcompsocitemizethanks{\IEEEcompsocthanksitem All authors are with Computer Vision Lab, ETH Zurich, Switzerland.}}

\IEEEtitleabstractindextext{%
\begin{abstract} \justifying
Due to the fact that fully supervised semantic segmentation methods require sufficient fully-labeled data to work well and can not generalize to unseen classes, few-shot segmentation has attracted lots of research attention. Previous arts extract features from support and query images, which are processed jointly before making predictions on query images. The whole process is based on convolutional neural networks (CNN), leading to the problem that only local information is used. In this paper, we propose a TRansformer-based Few-shot Semantic segmentation method (TRFS). Specifically, our model consists of two modules: Global Enhancement Module (GEM) and Local Enhancement Module (LEM). GEM adopts transformer blocks to exploit global information, while LEM utilizes conventional convolutions to exploit local information, across query and support features. Both GEM and LEM are complementary, helping to learn better feature representations for segmenting query images. Extensive experiments on PASCAL-5$^i$ and COCO datasets show that our approach achieves new state-of-the-art performance, demonstrating its effectiveness. Code and pretrained models will be available at \url{https://github.com/GuoleiSun/TRFS}.
 \end{abstract}
}


\maketitle
\IEEEdisplaynontitleabstractindextext

%
\IEEEpeerreviewmaketitle

\section{Introduction}
With the rapid development of deep learning, semantic segmentation, one of the most fundamental tasks in computer vision, has achieved significantly better performance than before~\cite{chen2017rethinking,chen2018encoder}. However, the fully supervised methods are largely limited by their dependence on sufficient datasets with pixel-wise ground-truth annotations, which requires intensive manual labor. Thus, numerous efforts~\cite{dai2015boxsup,papandreou2015weakly,lee2019ficklenet,wang2019panet,tian2020prior} are motivated to address this problem. Different tasks such as weakly supervised semantic segmentation, domain adaption in segmentation, and few-shot semantic segmentation, have been proposed. Among them, few-shot semantic segmentation is an appealing direction and has attracted much research attention \cite{shaban2017one,dong2018few,wang2019panet,tian2020prior,zhang2019canet,zhang2019pyramid}.

The goal of few-shot semantic segmentation is to segment a query image, given the support set which is comprised of a few support images and corresponding ground-truth masks. Many recent proposed few-shot segmentation approaches~\cite{wang2019panet,tian2020prior,zhang2019canet} follow this pipeline: first extract features for both support and query images, then process the support features and query features, and finally make predictions on query images based on the refined features. PL~\cite{dong2018few} and PANet~\cite{wang2019panet} learn prototypes for each class and compute cosine similarity between prototypes and features to make predictions. Another stream of works, including CANet~\cite{zhang2019canet}, PFENet~\cite{tian2020prior}, and PGNet~\cite{zhang2019pyramid}, adopt convolutional layers to process features.

Despite their success, they typically only use local information when processing features, either by pixel-wise similarity computation or convolutional layers, while global relationship modelling is of vital importance for scene understanding. Motivated by this, we propose to exploit the global information when processing support and query features. Recently, transformers have been proven to be effective in various vision tasks~\cite{dosovitskiy2021image,liu2021transformer,carion2020end,zheng2021rethinking,sun2021boosting} due to its ability in establishing long-range relationships within image features. Inspired by this, we propose transformer-based few-shot semantic segmentation (TRFS). Specifically, our method (shown in Fig.~\ref{fig:framework}) contains two modules: Global Enhancement Module (GEM) and Local Enhancement Module (LEM). The former refines features with global receptive fields while the latter focuses on local information. The combination of both modules provides better feature refinement for segmenting the query images, guided by the support set.

Our contributions are as follows. First, we address the value of global information in few-shot semantic segmentation, which is achieved by adopting transformers. Second, we show that global information and local information are complementary in few-shot semantic segmentation. The combination of both information performs better than individual one. Third, we achieve state-of-the-art results on two standard benchmarks.




\section{Related Work}
\subsection{Semantic Segmentation}
Semantic segmentation is one of the most fundamental tasks in the computer vision community. It aims to predict the semantic label for each pixel in a natural image. In the era of deep learning, it has achieved tremendous progress~\cite{chen2015semantic,chen2017deeplab,chen2017rethinking,chen2018encoder}, including popular approaches such as DeepLab~\cite{chen2018encoder}, DPN~\cite{liu2015semantic}, and CRF-RNN~\cite{zheng2015conditional}. 

Despite the fact that promising performance has been obtained on several standard large-scale benchmarks, those methods need sufficient well-labeled data (pixel-wise annotations) to work well. However, pixel-wise labeled data are very expensive, limiting many real-world applications of semantic segmentation. To relieve the problem, weakly supervised semantic segmentation and few-shot semantic segmentation have attracted more and more attention. Both directions are important, but they have different underlying goals, as well as formulations. Specifically, the former aims to replace the pixel-wise annotations to other weaker forms of annotations, such as bounding boxes~\cite{dai2015boxsup,papandreou2015weakly}, scribbles~\cite{lin2016scribblesup}, points~\cite{bearman2016s}, and image-level labels~\cite{ahn2018learning,lee2019ficklenet,sun2020mining,liu2020leveraging}. The latter focuses on improving model's generalizability to unseen classes with only a few well-labeled samples. In this work, we target on the few-shot setting.


\subsection{Few-shot Segmentation}
Few-shot semantic segmentation has gained lots of research interests after that Shaban \etal \cite{shaban2017one} first tackled this problem by proposing to adapt the classifier for each class, conditioned on the support set. One stream of works~\cite{dong2018few,wang2019panet} involves with learning prototypes. PL~\cite{dong2018few} learns prototypes for different classes and the prediction is made by computing the cosine similarity between the features and the prototypes. PANet~\cite{wang2019panet} makes progress by learning consistent prototypes through introducing alignment regularization. Another stream of methods~\cite{tian2020prior,zhang2019canet,zhang2019pyramid} concatenates the support and query features and let the network to figure out the relations between query and support, so that the segmentation can be conducted based on the clues given by the support set. 
As discussed before, existing methods merely use the local information within query-support features, while the global information is ignored. As we know, global relationship modelling is of vital importance for scene understanding in computer vision \cite{dosovitskiy2021image,liu2021transformer,wang2018non}. Motivated by this, this paper boosts the few-shot semantic segmentation by adopting transformers to exploit the global information over the merged query-support features.

\subsection{Transformer}
Recently, transformer, first introduced in natural language processing~\cite{vaswani2017attention,dai2019transformer,devlin2019bert,yang2019xlnet}, has attracted lots of research attention in the computer vision community. It relies on a multi-head self-attention (MHSA) module and a multi-layer perceptron (MLP), to model the global relationship within input sequences. For vision tasks, images or features are first converted into sequences of vectors, the global interactions within which are then modelled by the transformers. Since the pioneer works such as ViT~\cite{dosovitskiy2021image} and DETR~\cite{carion2020end}, it has been shown to be effective in various tasks, including image classification~\cite{yuan2021tokens,wang2021pyramid,liu2021transformer}, object detection~\cite{carion2020end}, semantic/instance segmentation~\cite{zheng2021rethinking,ding2021looking}, video segmentation~\cite{wang2020end}, crowd counting~\cite{sun2021boosting,liang2021transcrowd}, depth estimation~\cite{li2020revisiting,yang2021transformers}, domain adaptation~\cite{zhang2021detr,yang2021transformer}, and virtual try-on~\cite{ren2021cloth}. In particular, ViT~\cite{dosovitskiy2021image} divides the image into patches and converts them to sequences of features, which are then used as the input to the transformers. In contrast, DETR~\cite{carion2020end} directly exploits CNN features as the input for transformers for object detection. For a more complete survey for vision transformers, please refer to \cite{khan2021transformers}. However, to the best of our knowledge, there is no exploration of transformers for few-shot semantic segmentation. In this paper, we fill the gap and demonstrate the effectiveness of global relationship modelling using transformers in this task.


\begin{figure*}[t]
\centering
      \includegraphics[width=1.0\linewidth]{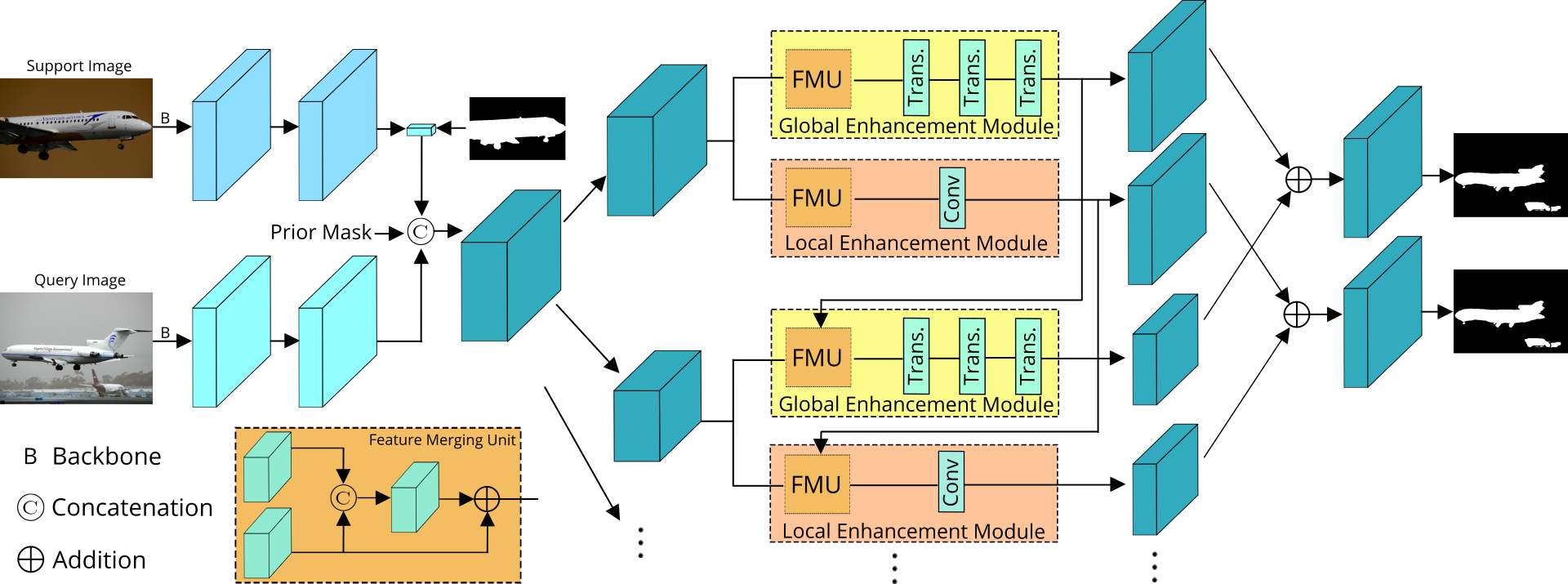}
\caption{Framework overview. The pipeline is shown for the one-shot case. The details of feature merging unit (FMU) are shown in bottom-left. Given a support image and the associated ground-truth mask, our framework segments the query image for the target class (plane in this example). The core of our model is combination of the Global Enhancement Module (GEM) and Local Enhancement Module (LEM). The former explores the global information, while the latter focuses on the local information. The synthesis of both modules enhances feature representations for few-shot semantic segmentation. Best viewed in color.}
\label{fig:framework}
\end{figure*}

\section{Methods}
In this section, we first give formal definition of few-shot semantic segmentation. Then we introduce how to fuse support and query features. Finally, we explain our global enhancement module, local enhancement module, and the loss functions for training our model.

\subsection{Problem Formulation}
For few-shot semantic segmentation, all classes are divided into two disjointed class set $\mathcal{C}_{train}$ (base classes) and $\mathcal{C}_{test}$ (unseen classes), where $\mathcal{C}_{train} \cap \mathcal{C}_{test}=\emptyset$. The goal of this task is to train the model on $\mathcal{C}_{train}$ and evaluate the model on unseen classes $\mathcal{C}_{test}$. Both training and testing are conducted in episodes~\cite{vinyals2016matching,shaban2017one}. In particular, let $\mathcal{D}_{train}=\{(I_i, M_i, \{I_{i}^{S_k}, M_{i}^{S_k}\}_{k=1}^{K})\}_{i=1}^{N_{tr}}$ denote $N_{tr}$ training episodes from $\mathcal{C}_{train}$. Here, $I_i$ and $M_i$ are $i^{th}$ query image and the corresponding ground-truth mask, which form a query set. Each query set is associated with a small ($K$-shot) support set $\{I_{i}^{S_k}, M_{i}^{S_k}\}_{k=1}^{K}$, where $I_{i}^{S_k}$ and $M_{i}^{S_k}$ are $k^{th}$ support image and the corresponding ground-truth mask for the $i^{th}$ query image. Each training episode (query and support sets) focuses on the same class, sampled from $\mathcal{C}_{train}$. For evaluation, let $\mathcal{D}_{test}=\{(I_i, \{I_{i}^{S_k}, M_{i}^{S_k}\}_{k=1}^{K})\}_{i=1}^{N_{te}}$ denote the test episodes. For each test episode, the model needs to segment $I_i$ based on the information given by the support set $\{I_{i}^{S_k}, M_{i}^{S_k}\}_{k=1}^{K}$, \ie, segment the same class as the ground-truth masks of the support set.

\subsection{Feature Fusion}
Here, we introduce how the input features for global and local enhancement modules are generated. Our formulation is based on a single episode of $\{(I, M, \{I^{S_k}, M^{S_k}\}_{k=1}^{K})\}$, for notation simplicity.
Following previous works~\cite{wang2019panet,tian2020prior}, we start from the query features and support features encoded by the ImageNet~\cite{russakovsky2015imagenet} pre-trained backbone, whose parameters are fixed throughout the training process. Let $\Theta$, $F_Q\in \mathcal{R}^{H \times W\times C}$, and $F_{S_k}\in \mathcal{R}^{H \times W\times C}$ denote the backbone function, the query feature map, and the $k^{th}$ support feature map, respectively. $C$, $H$, and $W$ are feature channel number, height, and width, respectively. We have
\begin{align}\small
\begin{split}
        &F_Q=\Theta(I),~~~F_{S_k}=\Theta(I^{S_k}). \\
\end{split}
\end{align}
%
As mentioned above, few-shot semantic segmentation aims to segment the query image based on the clues given by support images and support ground-truth masks. Hence, we obtain the support prototype $F_S$, which is used to guide the segmentation of the query image, as follows:
\begin{align}\small
\begin{split}
        &F_{S}=\frac{\sum_{k=1}^{K}GAP(F_{S_k}[M^{S_k},:])}{K}, \\
\end{split}
\end{align}
where $GAP$ denotes global average pooling across the spatial dimension, and the ground-truth mask $M^{S_k} \in \mathcal{R}^{H\times W}$ is already resized to the feature resolution. Intuitively, the support prototype $F_S \in \mathcal{R}^C$ is a feature vector averaged from the foreground features in the support set, which encodes the representative information of the target class. Different few-shot segmentation methods differ in the way to use the support prototype $F_S$ to guide the segmentation of query images. PANet~\cite{wang2019panet} directly computes similarity between query features and support prototypes. PFENet~\cite{tian2020prior} combines different features into a new feature, and then uses several convolutional layers to refine it. Since concatenating features enables further refinement and enhancement, we follow PFENet~\cite{tian2020prior} to generate input features $X\in \mathcal{R}^{H\times W \times (2C+1)}$ by concatenating query features ($\mathcal{R}^{H\times W \times C}$), expanded support prototypes ($\mathcal{R}^{H\times W \times C}$), and the prior mask ($\mathcal{R}^{H\times W \times 1}$), which will be used in global and local information modules. For details of computing the prior mask, we refer to PFENet~\cite{tian2020prior}. Simply, it pre-estimates the probability of pixels belonging to the target class, using high-level features.

\subsection{Global and Local Enhancement Modules}
\noindent\textbf{Multi-scale Processing.}
Inspired by the fact that object size in both support and query images can vary largely~\cite{tian2020prior}, we design a multi-scale framework with the input feature map $X$ so that information over different scales can be utilized, as shown in Fig.~\ref{fig:framework}. To obtain features in different scales, we use adaptive average pooling. Let $R=\{R^1, R^2, ..., R^n\}$ denote the spatial resolution after the average pooling, and we assume $R^1>R^2>...>R^n$. Feature $X_i$ with spatial size of $R^i$ can be obtained by 
\begin{align}\small
\begin{split}
        &X_i={\rm GAP}_{R^i}(X), \\
\end{split}
\end{align}
where ${\rm GAP}_{R^i}$ indicates adaptive average pooling so that the output feature has the size of $R^i$. Hence, a feature pyramid of $\{X_1, X_2,...,X_n\}$ is obtained, each of which will be processed by both global enhancement module and local enhancement module. In particular, $X_i \in \mathcal{R}^{R_i\times R_i \times (2C+1)}$.

\myPara{Global Enhancement Module (GEM).}
Different from previous study~\cite{tian2020prior} which uses convolutional layers to refine the combined features, we propose to adopt transformers to enhance the features so that global information can be exploited, as shown in Fig.~\ref{fig:framework}. We first reduce the channel dimension of $X_i$ by a fully connected layer and obtain $X_i^{'}\in \mathcal{R}^{R_i\times R_i \times C}$. $X_i^{'}$ first goes through the \textbf{Feature Merging Unit (FMU)}, which merges the output feature from the branch refining $X_{i-1}$. If $i=1$, then no feature merging is performed and $X_i^{'}$ is directly output from FMU. Let $Y_i \in \mathcal{R}^{R_i \times R_i \times C}$ denote the output of FMU, given by
\begin{align}\small
\begin{split}
        Y_i= 
            \begin{cases}
                {\rm Conv}_{1\times 1}({\rm Concat}(X_i^{'},T_{i-1}^{L}))+X_i^{'},& \text{if } x> 1\\
                X_i^{'},              & \text{if } x= 1
            \end{cases}
\end{split}
\end{align}
where ${\rm Concat}(\cdot)$ denotes feature concatenation across channels, ${\rm Conv}_{1\times 1}$ represents $1\times 1$ convolution with output channel of $C$ and interpolation is not shown for simplicity.

We reshape $Y_i$ into $\mathcal{R}^{{R_i}^2 \times C}$. Then, the obtained sequence of vectors is processed by $L$ transformer blocks to explore the global information, denoted as follows:
\begin{align}\small
\begin{split}
        &T_i^{0}=Y_i,\\
        &\hat{T}_{i}^{l}={\rm MHSA}(T_i^{l-1})+T_i^{l-1},  ~~~l=1,...,L,\\
        &T_{i}^{l}={\rm MLP}(\hat{T}_{i}^{l})+\hat{T}_{i}^{l}, ~~~l=1,...,L,\\
\end{split}
\end{align}
in which ${\rm MHSA}(\cdot)$ denotes the standard multi-head self-attention
in transformer \cite{vaswani2017attention,dosovitskiy2021image},
and ${\rm MLP}(\cdot)$ is a two-layer multi-layer perceptron.
After the transformers, we obtain $T_{i}^{L} \in\mathcal{R}^{{R_i}^2 \times C}$, which is reshaped back to $\mathcal{R}^{R_i \times R_i \times C}$. In our experiments, we use $L=3$. After processing different scales, we have $\{T_{1}^{L}, T_{2}^{L},...,T_{n}^{L}\}$. The final output feature from the global enhancement module is formed by interpolation and concatenation of $n$ enhanced feature maps $T_{i}^{L}$, denoted as 
\begin{align}\small
\begin{split}
        &T_i^{0}=Y_i,\\
        &\hat{T}_{i}^{l}={\rm MHSA}(T_i^{l-1})+T_i^{l-1},  ~~~l=1,...,L,\\
        &T={\rm Concat}(T_{1}^{L}, T_{2}^{L},...,T_{n}^{L}),\\
\end{split}
\end{align}
where ${\rm Concat}(\cdot)$ indicates feature concatenation across the channel dimension, and interpolation is not shown for simplicity. $T$ is used to predict the target mask $M$.

\myPara{Local Enhancement Module (LEM).} 
The local enhancement module follows the same pipeline as GEM. $X_i$ is processed by a fully connected layer and FMU to generate $Y_i$. Different from GEM which utilizes transformer blocks to process $Y_i$, LEM exploits conventional convolution to refine $Y_i$, in order to encode the local information. Both global and local information can be complementary. After LEM, let $\{Z_{1}, Z_{2},...,Z_{n}\}$ denote the output features from different scales. Similarly, the final output feature from LEM is formed by the interpolation and concatenation of $Z_{i}$, denoted as
\begin{align}\small
\begin{split}
        &Z={\rm Concat}(Z_{1}, T_{2},...,T_{n}),\\
\end{split}
\end{align}
where interpolation is omitted for simplicity. $Z$ is used to predict the target mask $M$.

\subsection{Loss Functions}
Both the features from GEM and LEM are used to predict the target mask of the query image, whose losses are $\mathcal{L}_{\rm GEM}$ and $\mathcal{L}_{\rm LEM}$. The final loss for the whole network is defined as
\begin{align}\small
\begin{split}
        &\mathcal{L}=\mathcal{L}_{\rm GEM}+\mathcal{L}_{\rm LEM}\\
\end{split}
\end{align}
Here, both $\mathcal{L}_{\rm GEM}$ and $\mathcal{L}_{\rm LEM}$ are 
common cross-entropy loss for semantic segmentation
\cite{chen2015semantic,chen2017deeplab,chen2017rethinking}.
During testing, the final prediction for query image is the average of the predictions output from the global enhancement module and the local enhancement module.

\section{Experiments}
Experiments are conducted on two benchmark few-shot semantic segmentation datasets~\cite{shaban2017one,lin2014microsoft} to validate the effectiveness of the proposed approach. We begin this section by 
introducing our experimental setting, followed by state-of-the-art comparisons with previous methods. Finally, we show ablation studies to examine the effectiveness of key components of our model.

\subsection{Experimental Setup}
\noindent\textbf{Datasets.} We conduct experiments on standard benchmarks of PASCAL-5$^{i}$~\cite{shaban2017one} and COCO~\cite{lin2014microsoft} to evaluate the proposed method. The PASCAL-5$^{i}$ is constructed from PASCAL VOC 2012~\cite{everingham2010pascal} and SDS~\cite{hariharan2014simultaneous} datasets. It has 20 classes, which are evenly split into 4 groups, with 5 classes for each. The COCO is a more challenging dataset, having 82,783 training images and 40504 test images. The whole 80 classes are evenly divided into 4 folds, with 20 classes for each. For both datasets, the split of class groups follows previous works~\cite{wang2019panet,tian2020prior} for fair comparisons. The evaluation is done by cross-validation. Specifically, for each split, three groups of classes are used as base classes, while the remaining one is used as unseen classes. For test,  we randomly sample 5,000 query-support pairs for each fold following PFENet~\cite{tian2020prior,shaban2017one} on PASCAL-5$^{i}$ dataset. Since COCO has a large validation set, we sample 20,000 query-support pairs on each fold during the evaluation.

\myPara{Implementation Details
.} Following previous arts~\cite{wang2019panet,tian2020prior}, we test our method on ImageNet~\cite{russakovsky2015imagenet} pre-trained backbones: VGG-16~\cite{simonyan2015very}, ResNet-50~\cite{he2016deep} and ResNet-101~\cite{he2016deep}. For transformer parameters, we set number of heads in MHSA as 8 and MLP ratio as 4. GELU non-linear activation and Layernorm are used in transformer layers. For dataloader, we follow the official implementation of PFENet~\cite{tian2020prior}. Specifically, data augmentations of horizontal flip, random rotation within 10 degrees, and random cropping of $473\times 473$ are used. For optimizing the network, we use SGD optimizer with momentum and weight decay set to 0.9 and 0.0001, respectively. `Poly' learning rate scheduler is used with power parameter set as 0.9. For PASCAL-5$^{i}$ dataset, the model is trained for 200 epochs with batch size of 4 on single GPU. For COCO dataset, the model is trained for 50 epochs with batch size of 32 on 4 GPUs. Our framework is implemented in PyTorch. The code and trained models will be released.

\myPara{Evaluation Metrics.} Following previous works~\cite{wang2020few,wang2019panet,tian2020prior}, we report mean intersection over union (mIoU) on individual folds and the final averaged mIoU on all folds. Note that our results are all single-scale results without any post-processing such as multi-scale testing or DenseCRF~\cite{krahenbuhl2011efficient}.

\subsection{Comparison with State-of-the-Arts}
The state-of-the-art comparisons for PASCAL-5$^{i}$ and COCO datasets are shown in Table~\ref{table:results_pascal} and Table~\ref{table:results_coco}, respectively.

From the results, we have four observations. First, the proposed method achieves better performance than the compared state-of-the-art approaches, which demonstrates the effectiveness of global and local enhancement module. Specifically, our method outperforms PFENet~\cite{tian2020prior} by 1.3\% in terms of mean mIoU over 4 folds in 5-shot setting on PASCAL-5$^{i}$. For specific folds, the proposed approach achieves 3.0\% mIoU gain on fold-3 on 5-shot setting using ResNet-50. Since previous works~\cite{zhang2019canet,tian2020prior} only consider the local information to refine the merged query-support features while ours also takes global information into account, the performance gain of ours over those methods is attributed to the GEM. Second, our approach is robust across different backbones: VGG-16, ResNet-50 and ResNet-101. For these backbones, our method achieves consistent performance gain over the corresponding state-of-the-art methods. For instance, under 1-shot setting, the performance gain for VGG-16 and ResNet-50 are 1.0\% and 1.1\%, respectively. Third, the performance gain of our method over other methods is consistent for both 1-shot and 5-shot setting. Fourth, on the challenging COCO dataset, the proposed approach also obtains promising results. Specifically, our method (VGG-16) achieves 1.9\% gain over the current state-of-the-art method PFENet in terms of mean mIoU under 5-shot setting.

\begin{figure*}[t]
  \centering
      \includegraphics[width=0.99\linewidth]{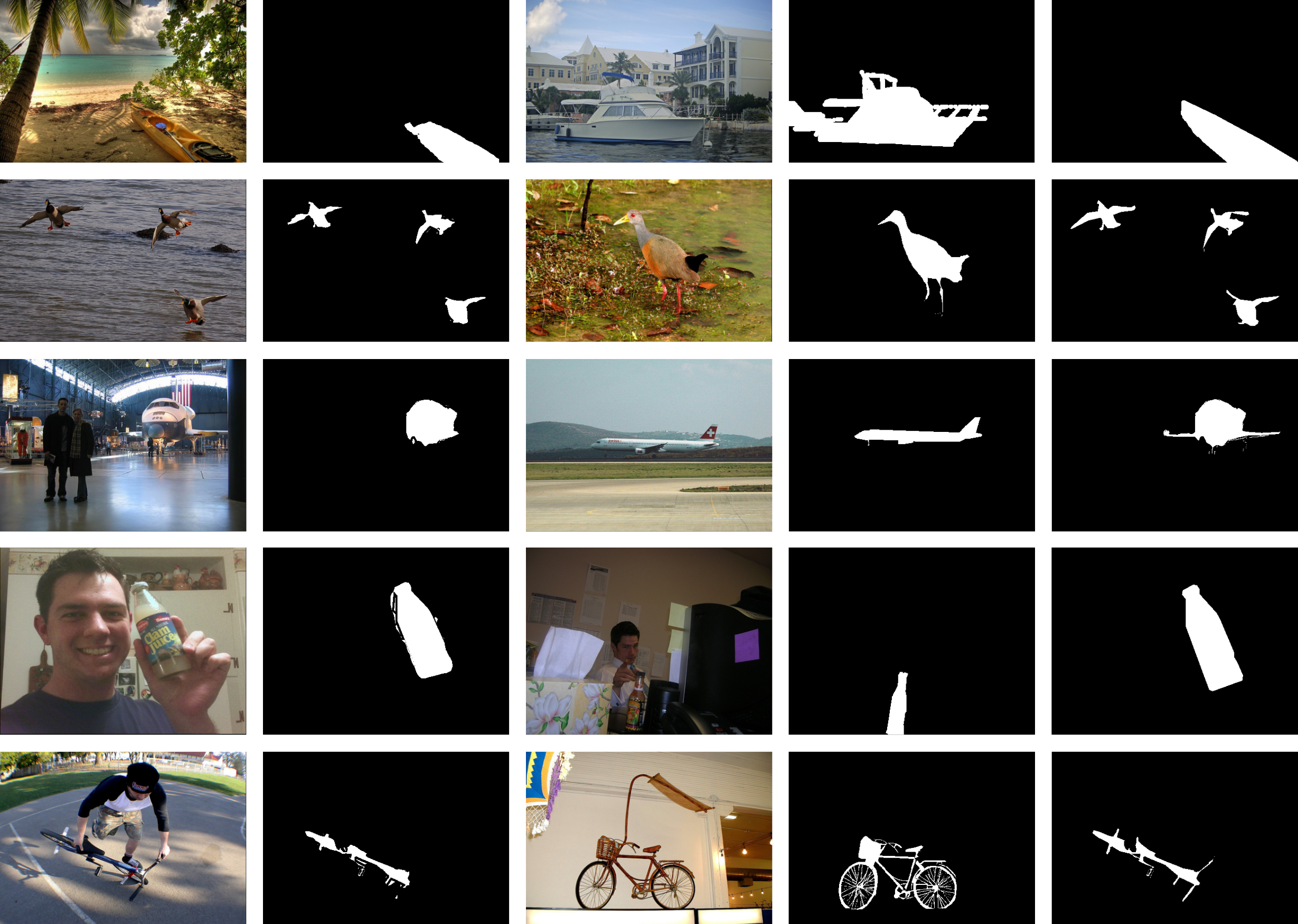}
\caption{Qualitative results on novel/unseen classes on PASCAL-5$^{i}$ dataset using ResNet-50 model. From \textit{left} to \textit{right}: query image, query prediction, support image, support ground-truth mask, and query ground-truth mask. It shows that our method performs well on novel classes in 1-shot setting where a single support image and its associated mask are given to guide the segmentation.}
\label{fig:qual}
\end{figure*}

Qualitative results on novel classes are shown in Fig.~\ref{fig:qual}. Our method performs well on 1-shot setting where only single support image and its ground-truth mask are given. The shown examples are challenging due to: the query or support images are unclear due to bad weather or shadow, objects in query and support images cover very different object regions, object size in query and support image varies significantly, or complicated/complex background exists.

\begin{table*}[!t]
\centering
\caption{Comparison with state-of-the-art methods on PASCAL-5$^{i}$ dataset. It shows that our method achieves new state-of-the-art performance on this dataset.}
\label{table:results_pascal}
\resizebox{0.999\textwidth}{!}{%
\begin{tabular}{c|c|c|c|c|c|c|c|c|c|c|c}
\hline\thickhline
\multirow{2}{*}{Methods} & \multirow{2}{*}{Publication} & \multicolumn{5}{c|}{1-Shot}              & \multicolumn{5}{c}{5-Shot}              \\ \cline{3-12} 
                         &                              & Fold-0 & Fold-1 & Fold-2 & Fold-3 & Mean & Fold-0 & Fold-1 & Fold-2 & Fold-3 & Mean \\ \hline
\multicolumn{12}{c}{VGG-16 Backbone}                                                                                                        \\ \hline
OSLSM~\cite{shaban2017one}                    & BMVC17                       & 33.6   & 55.3   & 40.9   & 33.5   & 40.8 & 35.9   & 58.1   & 42.7   & 39.1   & 44.0 \\ \hline
co-FCN~\cite{rakelly2018conditional}                   & ICLRW18                      & 36.7   & 50.6   & 44.9   & 32.4   & 41.1 & 37.5   & 50.0   & 44.1   & 33.9   & 41.4 \\ \hline
SG-One~\cite{zhang2020sg}                   & TCYB20                       & 40.2   & 58.4   & 48.4   & 38.4   & 46.3 & 41.9   & 58.6   & 48.6   & 39.4   & 47.1 \\ \hline
AMP~\cite{siam2019adaptive}                      & ICCV19                       & 41.9   & 50.2   & 46.7   & 34.7   & 43.4 & 41.8   & 55.5   & 50.3   & 39.9   & 46.9 \\ \hline
PANet~\cite{wang2019panet}                    & ICCV19                       & 42.3   & 58.0   & 51.1   & 41.2   & 48.1 & 51.8   & 64.6   & \textbf{59.8}   & 46.5   & 55.7 \\ \hline
FWBF~\cite{nguyen2019feature}                     & \multicolumn{1}{c|}{ICCV19}  & 47.0   & 59.6   & 52.6   & 48.3   & 51.9 & 50.9   & 62.9   & 56.5   & 50.1   & 55.1 \\ \hline
RPMMs~\cite{yang2020prototype}                     & \multicolumn{1}{c|}{ECCV20}  &  47.1  & 65.8   & 50.6   &  48.5  & 53.0 & 50.0   &  66.5  &  51.9  & 47.6  & 54.0 \\ \hline
CRNet~\cite{liu2020crnet}                     & \multicolumn{1}{c|}{CVPR20}  &  -  &  -  &  -  &  -  & 55.2 &  -  &  -  &  -  &  - & 58.5 \\ \hline
PFENet~\cite{tian2020prior}                   & \multicolumn{1}{c|}{TPAMI20} & 56.9   & 68.2   & 54.4   & 52.4   & 58.0 & \textbf{59.0}   & 69.1   & 54.8   & 52.9   & 59.0 \\ \hline
Ours                     & \multicolumn{1}{c|}{-}       &    \textbf{58.8}    &  \textbf{68.4}      &     \textbf{54.8}   &    \textbf{53.8}    & \textbf{59.0}     &    57.8    &   \textbf{69.4}     &     54.8   &    \textbf{56.4}    & \textbf{59.6}     \\ \hline
\multicolumn{12}{c}{ResNet-50 Backbone}                                                                                                     \\ \hline
CANet~\cite{zhang2019canet}                    & \multicolumn{1}{c|}{CVPR19}  & 52.5   & 65.9   & 51.3   & 51.9   & 55.4 & 55.5   & 67.8   & 51.9   & 53.2   & 57.1 \\ \hline
PGNet~\cite{zhang2019pyramid}                    & \multicolumn{1}{c|}{ICCV19}  & 56.0   & 66.9   & 50.6   & 50.4   & 56.0 & 54.9   & 67.4   & 51.8   & 53.0   & 56.8 \\ \hline
RPMMs~\cite{yang2020prototype}                     & \multicolumn{1}{c|}{ECCV20}  &  55.2  & 66.9   & 52.6   & 50.7   & 56.3 &  56.3  & 67.3  & 54.5   & 51.0  & 57.3 \\ \hline
CRNet~\cite{liu2020crnet}                     & \multicolumn{1}{c|}{CVPR20}  &  -  &  -  &  -  &  -  & 55.7 &  -  &  -  &  -  &  - & 58.8 \\ \hline
PFENet~\cite{tian2020prior}                   & \multicolumn{1}{c|}{TPAMI20} & 61.7   & 69.5   & 55.4   & 56.3   & 60.8 & 63.1   & 70.7   & \textbf{55.8}   & 57.9   & 61.9 \\ \hline
Ours                     & \multicolumn{1}{c|}{-}       &    \textbf{62.9}    &  \textbf{70.7}      &     \textbf{56.5}   &    \textbf{57.5}    &   \textbf{61.9}   &   \textbf{65.0}     & \textbf{71.2}       &  55.5      &     \textbf{60.9}   &    \textbf{63.2}  \\ \hline
\end{tabular}
}
\end{table*}

\begin{table*}[!t]
\centering
\caption{Comparison with state-of-the-art methods on COCO dataset. It shows that our method achieves new state-of-the-art performance on this dataset.}
\label{table:results_coco}
\resizebox{0.999\textwidth}{!}{%
\begin{tabular}{c|c|c|c|c|c|c|c|c|c|c|c}
\hline\thickhline
\multirow{2}{*}{Methods} & \multirow{2}{*}{Publication}  & \multicolumn{5}{c|}{1-Shot}              & \multicolumn{5}{c}{5-Shot}              \\ \cline{3-12} 
                         &                    & Fold-0 & Fold-1 & Fold-2 & Fold-3 & Mean & Fold-0 & Fold-1 & Fold-2 & Fold-3 & Mean \\ \hline
\multicolumn{12}{c}{VGG-16 Backbone}                                                                                                        \\ \hline
PANet~\cite{wang2019panet}    & ICCV19    & -   &  -  &  -  &  -  & 20.9 &  -  &  -  & -   &  -  & 29.7 \\ \hline
FWBF~\cite{nguyen2019feature} & \multicolumn{1}{c|}{ICCV19}  &  18.4  &  16.7  & 19.6   & 25.4   & 20.0 &  20.9  &  19.2  & 21.9   & 28.4   & 22.6 \\ \hline
PFENet~\cite{tian2020prior}      & \multicolumn{1}{c|}{TPAMI20}    &  {33.4}  & 36.0   & 34.1   & {32.8}   & 34.1 &  {35.9}  & 40.7   & {38.1}   & 36.1   & 37.7 \\ \hline
Ours                     & \multicolumn{1}{c|}{-}     &    \textbf{34.2}    &  \textbf{38.8}      & \textbf{35.3}       &     \textbf{33.3}   &    \textbf{35.4}  &     \textbf{37.8}   &    \textbf{43.8}    & \textbf{39.7}       &  \textbf{36.9}      &     \textbf{39.6} \\ \hline

\multicolumn{12}{c}{ResNet-101 Backbone}                                                                                                     \\ \hline
FWBF~\cite{nguyen2019feature} & \multicolumn{1}{c|}{ICCV19}  & 19.9   & 18.0   & 21.0   & 28.9   & 21.2 & 19.1   & 21.5   & 23.9   & 30.1   & 23.7 \\ \hline
DAN~\cite{wang2020few}       & \multicolumn{1}{c|}{ECCV20} &  -  &  -  &  -  &  -  & 24.4 &  -  & -   &   - & -  & 29.6 \\ \hline

PFENet~\cite{tian2020prior}                   & \multicolumn{1}{c|}{TPAMI20}  &  \textbf{34.3}  & 33.0   & 32.3   & 30.1   & 32.4 &  \textbf{38.5}  & 38.6   &  38.2  & 34.3  & 37.4 \\ \hline
Ours                     & \multicolumn{1}{c|}{-}      &    31.8    &   \textbf{34.9}     &  \textbf{36.4}      & \textbf{31.4}       &    \textbf{33.6}  &     35.4   &    \textbf{41.7}    &  \textbf{42.3}      & \textbf{36.1}       & \textbf{38.9}     \\ \hline
\end{tabular}
}
\end{table*}

\subsection{Ablation Study}
We conduct ablation study on PASCAL-5$^{i}$ dataset to validate the contributions of the key components of our method. We also examine the effect of the number ($L$) of transformer layers and number of scales on the performance of our model.

\myPara{GEM and LEM.} We show the results of only using global enhancement module or local enhancement module in Table~\ref{table:ablation_study}. It shows that GEM and LEM perform similarly in terms of the final averaged mIoU over all folds, which achieve 60.6 and 60.8 respectively. However, the model using only GEM and the one using only LEM can have very different performance for a specific split. For example, using only GEM has mIoU of 70.9 while using only LEM achieves mIoU of 69.9 in fold-1. For fold-3, using LEM outperforms the model using GEM by 1.5\%. This suggests that GEM and LEM capture different information. When combining both GEM and LEM (our final model), we obtain better performance, demonstrating that GEM and LEM are complementary. 

 \begin{table}[!t]
	\centering	%
	\caption{Ablation study on the key components and $L$ (the number of transformer blocks)  on PASCAL-5$^{i}$ dataset using ResNet-50.}
	\label{table:ablation_study}
	\resizebox{0.49\textwidth}{!}{
	\begin{tabular}{c|c|c|c|c|c|c}
\hline
\multirow{2}{*}{} & \multirow{2}{*}{L} & \multicolumn{5}{c}{1-Shot}              \\ \cline{3-7} 
                  &                    & Fold-0 & Fold-1 & Fold-2 & Fold-3 & Mean \\ \hline
+GEM              & 3                  &   60.4    &  70.9      &  56.3      &     54.9   &    60.6  \\ \hline
+LEM              & 3                  &   60.9     &  69.9      &  56.1      &     56.4   &    60.8  \\ \hline
+GEM+LEM          & 3                  &    \textbf{62.9}    &  \textbf{70.7}      &  \textbf{56.5}      &     57.5   &    \textbf{61.9}  \\ \hline
+GEM+LEM          & 2                  &    62.1    &   70.6     &  54.1      &     \textbf{58.3}   &    61.3  \\ \hline
+GEM+LEM          & 4                  &     62.3   &   71.0     &  55.3      &     58.0  &     61.6 \\ \hline
\end{tabular}
}
\end{table}

\myPara{Number of Transformer Blocks.} 
We also evaluate the effect of different number of transformer blocks. It shows that our approach is robust to different choice of $L$, achieving comparable results. However, the best performance is observed when setting $L$ to be 3. It may due to that when using small $L=2$, global information is not fully explored. When using more transformer blocks ($L=4$), the network may overfit to the base classes and achieve a little worse performance on novel classes. 

 \begin{table}[!t]
	\centering	%
	\caption{Ablation study on different scale combinations on PASCAL-5$^{i}$ dataset using ResNet-50.}
	\label{table:ablation_study2}
	\resizebox{0.49\textwidth}{!}{
	\begin{tabular}{c|c|c|c|c|c}
\hline
\multirow{2}{*}{Scales}  & \multicolumn{5}{c}{1-Shot}              \\ \cline{2-6} 
                                     & Fold-0 & Fold-1 & Fold-2 & Fold-3 & Mean \\ \hline
[60]                                &    59.4   &  68.3      &  54.2      & 52.4       &    58.6  \\ \hline
[60,30]                                &   60.4     &  70.1      & 54.7  & 56.3   &    60.4  \\ \hline
[60,30,15]                 &    61.4    & 70.4       & 54.3      & \textbf{57.7}       & 61.0     \\ \hline
[60,30,15,8]      &   \textbf{62.9}     &   \textbf{70.7}     &    \textbf{56.5}    &   57.5     &  \textbf{61.9}    \\ \hline
\end{tabular}
}
\end{table}

\noindent\textbf{Number of Scales.}
We conduct multi-scale processing on input feature $X \in \mathcal{R}^{H\times W \times (2C+1)}$, which is the concatenation of  query features ($\mathcal{R}^{H\times W \times C}$), expanded support prototypes ($\mathcal{R}^{H\times W \times C}$), and the prior mask ($\mathcal{R}^{H\times W \times 1}$). Specifically, a series of adaptive average pooling operations with output scales $R=\{R^1, R^2, ..., R^n\}$ on $X$ is used. We ablate the effect of different scale variations in Table~\ref{table:ablation_study2}. In all our experiments, the input size is $473\times 473$. After going through the backbone, the height/weight ($H/W$) is 60. Hence, we start from a single scale of 60, and gradually add smaller scales of 30, 15, and 8. It shows that our method performs better when more scales are used. 
For all our results in the paper, we use $R=\{60,30,15,8\}$.

\section{Conclusion}
In this paper, we study the value of global information in few-shot semantic segmentation. We propose global enhancement module (GEM) to refine the query-support features, together with local enhancement module (LEM). GEM exploits global information via transformer layers while LEM utilizes local information through convolutional layers. The combination of both modules help to learn better features for segmenting query images. Our experiments show that GEM and LEM are complimentary, and the proposed method combining both GEM and LEM achieves state-of-the-art performance on two standard benchmark datasets, \ie, PASCAL-5$^{i}$ and COCO. Our qualitative results on novel classes show that our method provides promising segmentation masks on query images under challenging situations.

For future research, it is interesting to see if the feature interaction between global enhancement module and local enhancement module in intermediate layers can further boost the performance. It is also interesting to study the effect of other newly developed transformer layers in few-shot semantic segmentation.

\bibliographystyle{IEEEtran}
\bibliography{egbib}

\end{document}